\documentclass[10pt,letterpaper]{article}

\usepackage{cogsci}

\cogscifinalcopy 

\usepackage{pslatex}
\usepackage{hyperref}
\usepackage{apacite}
\usepackage{fancyhdr}
\usepackage{algorithm}
\usepackage{algpseudocode}
\usepackage{amsmath}
\usepackage{graphicx}

\usepackage{float} 

\title{CM-DQN: A Value-Based Deep Reinforcement Learning Model to Simulate Confirmation Bias}
 
\author{
  \begin{tabular}[t]{c@{\hspace{2cm}}c}
    {\large \bf Jiacheng Shen} & {\large \bf Lihan Feng} \\
    \small New York University Shanghai & \small New York University Shanghai \\
    \small Shanghai, China & \small Shanghai, China \\
    \small shen.patrick.jiacheng@nyu.edu & \small lf2383@nyu.edu
  \end{tabular}
}
  
\begin{document}
\maketitle
\begin{abstract}
In human decision-making tasks, individuals learn through trials and prediction errors. When individuals learn the task, some are more influenced by good outcomes, while others weigh bad outcomes more heavily. \cite{RSTD} Such confirmation bias can lead to different learning effects. In this study, we propose a new algorithm in Deep Reinforcement Learning, \textbf{CM-DQN}, which applies the idea of different update strategies for positive or negative prediction errors, to simulate the human decision-making process when the task's states are continuous while the actions are discrete. We test \textbf{CM-DQN} in Lunar Lander environment with confirmatory, disconfirmatory bias and non-bias to observe the learning effects. Moreover, we apply the confirmation model in multi-armed bandit problem (environment in discrete states and discrete actions), which utilizes the same idea with our proposed algorithm, as a contrast experiment to algorithmically simulate the impact of different confirmation bias in decision-making process. In both experiments, confirmatory bias indicates a better learning effect.
\end{abstract}

\section{Introduction}

Confirmation bias is a cognitive phenomenon where individuals tend to favor information that confirms their existing beliefs or hypotheses \cite{palminteri2017confirmation}. This bias can significantly impact decision-making processes and lead to unexpected outcomes. For instance, in the financial market, an investor might selectively seek out and interpret market information that supports their preconceived notions about a particular stock, thereby overlooking critical negative indicators. The significance of understanding this cognitive bias extends beyond individual decision-making. It plays a crucial role in areas such as politics, where confirmation bias can polarize debates and hinder consensus, or in science, where it can lead to preferential treatment of data and skew research findings.

Reinforcement learning shows its efficacy in modeling decision-making and becomes superior to human intelligence in game playing, and autonomous driving \cite{hester2018deep}. Studying confirmation bias in the context of reinforcement learning is significant in understanding the human decision-making process. As confirmation bias is a pervasive aspect of human cognition that influences human decision-making process, we can numerically analyze how decisions can be influenced by human pre-existing beliefs. Existed work models confirmation bias in multi-armed bandit problems by assigning different updating rates on value functions based on prediction error \cite{lefebvre2022normative}. However, our world is always continuous. Neural Network has emerged as a powerful universal approximator to approximate high-dimensional and continuous functions. Therefore, deep learning provides us a new perspective to study continuous confirmation bias.

In this project, we first studied the confirmation bias in the multi-arm bandit problem. Furthermore, to explore confirmation bias in the continuous decision process, we integrate the confirmation model with Deep Q Network.
In summary, we have the following contributions in our project:
\begin{enumerate}
    \item We studied the confirmation model in the context of the multi-armed bandit problem
    \item We proposed a new deep reinforcement learning algorithm with a confirmation model that solves continuous decision-making process problems.
    \item We compared the different types of bias in the confirmation model by numerical experiments.
\end{enumerate}
\section{Related Work}
\subsubsection{Confirmation Bias}
The term ‘confirmation bias’ has been used to refer to various distinct ways in which beliefs and expectations can influence the selection, retention, and evaluation of evidence (Klayman 1995; Nickerson 1998). Hahn and Harris (2014) offer a list of them including four types of cognitions: (1) hypothesis-determined information seeking and interpretation, (2) failures to pursue a falsificationist strategy in contexts of conditional reasoning, (3) a resistance to change a belief or opinion once formed, and (4) overconfidence or an illusion of validity of one’s own view. \cite{confirmation} In reinforcement learning-based decision-making simulation, the environment is unknown in most cases. Therefore, in our study, we mainly focus on the last 3 types of bias. The last 3 types of bias can be summarized into 2 types: confirmatory bias--people are more willing to choose the one that they believe is good, and disconfirmatory bias--people are less likely to choose the one that they have a bad impression.
\subsubsection{Risk-Sensitive Temporal Difference (RSTD) Model with separate learning rates}
Risk-sensitive Temporal Difference (RSTD) model combines the concepts of time-difference learning and risk perception for decision making and learning under uncertain environment. By modeling the uncertainty of the environment as a probability distribution and taking into account the risk preference of decision makers, the model enables individuals to adapt more flexibly to different risk scenarios. The research from Rosenbaum, Grassie, and Hartley \cite{RSTD} applies the RSTD model with separate learning rates for better-than-expected ($\alpha^+$) and worse-than-expected ($\alpha^-$) outcomes, whose purpose is to index the valence bias in learning when doing a risk-sensitive decision-making RL task.
allowing us to index valence biases in learning. 
\subsubsection{Bayesian learning in modeling psychological bias}
Zimper and Ludwig previously developed formal models of Bayesian learning with psychological bias as alternatives to rational Bayesian learning based on Choquet expected utility theory. They introduced parameters, one is to measure the lack of confidence (ambiguity) of the decision maker in his additive prior belief, and the second parameter to measure the degree of optimism, respectively pessimism, that the decision maker attaches to a resolution of ambiguity in the course of the learning process. They proposed an alternate model to quantize the psychological bias when making decisions. \cite{baye}
\subsubsection{Deep Q Network}
In recent years, there has been significant interest in the development of reinforcement learning algorithms capable of learning directly from high-dimensional sensory input, such as images or raw sensor data. One notable algorithm that has emerged in this domain is the Deep Q-Network (DQN) algorithm \cite{hester2018deep}. DQN combines deep neural networks with Q-learning, a classical reinforcement learning technique, to learn value functions directly from raw pixel inputs. The core idea behind DQN is to approximate the optimal action-value function $Q(s,a)$ which represents the expected cumulative reward of taking action $a$ in state $s$, using a deep neural network parameterized by $Q(s,a;\theta)$. By iteratively updating the network parameters to minimize the temporal difference error between the current estimate and the target value, DQN is able to learn effective policies for a wide range of tasks. One key advantage of DQN is its ability to handle high-dimensional state spaces, making it well-suited for tasks where traditional tabular methods are infeasible. Furthermore, DQN introduces experience replay and target networks to stabilize learning and improve sample efficiency, respectively. Despite its successes, DQN and its variants are not without limitations. For example, they often require large amounts of data and computation to learn effectively, and they may struggle in environments with sparse rewards or complex dynamics. Nonetheless, DQN has served as a foundational model in the field of deep reinforcement learning and has inspired numerous extensions and improvements. In the context of this study, we draw upon the principles of DQN to develop a novel algorithm capable of learning in continuous state spaces and addressing specific challenges related to confirmation bias in reinforcement learning tasks.
\section{Method}
\subsection{Preliminary}
\subsubsection{Markov Decision Processes (MDPs)}

A \textbf{Markov Decision Process (MDP)} provides a mathematical framework for modeling decision-making in situations where outcomes are partly random and partly under the control of a decision-maker. MDPs are widely used in optimization, control theory, artificial intelligence, machine learning, economics, and more.

An MDP is defined by a tuple $\langle S, A, P, R, \gamma \rangle$, where:

\begin{itemize}
\item $\mathcal{S}$ is a finite set of states.
\item $\mathcal{A}$ is a finite set of actions.
\item $P$ is a state transition probability matrix, $P_{ss'}^a = P(S_{t+1} = s' | S_t = s, A_t = a)$.
\item $R$ is a reward function, $R_s^a = E[R_{t+1} | S_t = s, A_t = a]$.
\item $\gamma$ is a discount factor, $\gamma \in [0, 1]$.
\end{itemize}

\subsubsection{Bellman Equation}

The \textbf{Bellman equation}, named after Richard Bellman \cite{barron1989bellman}, is a necessary condition for optimality associated with the mathematical optimization method known as dynamic programming. It writes the value of a decision problem at a certain point in time in terms of the payoff from some initial choices and the value of the remaining decision problem that results from those initial choices. This breaks a dynamic optimization problem into a sequence of simpler subproblems, as Bellman's Principle of Optimality prescribes.

For a policy $\pi$, the Bellman equation is:

\[ V^\pi(s) = \sum_{a \in A} \pi(a|s) \left( R_s^a + \gamma \sum_{s' \in S} P_{ss'}^a V^\pi(s') \right) \]

The optimal state-value function satisfies the Bellman optimality equation:

\[ V^*(s) = \max_{a \in A} \left( R_s^a + \gamma \sum_{s' \in S} P_{ss'}^a V^*(s') \right) \]

\subsubsection{Q-Learning}

\textbf{Q-learning} is a model-free reinforcement learning algorithm \cite{watkins1992q}. The goal of Q-learning is to learn a policy, which tells an agent what action to take under what circumstances. It does not require a model (hence the connotation "model-free") of the environment, and it can handle problems with stochastic transitions and rewards, without requiring adaptations.

For any finite MDP, Q-learning finds an optimal policy in the sense of maximizing the expected value of the total reward over any and all successive steps, starting from the current state. Q-learning can identify an optimal action-selection policy for any given (finite) MDP, given infinite exploration time and a partly random policy.

The Q-learning algorithm uses a function Q that is similar to the value function in the Bellman equation. The Q function takes two arguments: the current state s and an action a. The Q function returns the expected future reward of that action at that state. This function can be estimated using temporal difference learning.

The Q-learning update rule is as follows:

\[ Q(s, a) \leftarrow Q(s, a) + \alpha \left( r + \gamma \max_{a'} Q(s', a') - Q(s, a) \right) \]

where:
\begin{itemize}
\item $\alpha$ is the learning rate.
\item $r$ is the reward for taking action $a$ in state $s$.
\item $\gamma$ is the discount factor.
\end{itemize}

\subsubsection{Confirmation model in multi-armed bandit problem}
In Lefebvre's work \cite{lefebvre2022normative}, they denote prediction error under choosing bandit $i$ as $\delta^i$ where \begin{equation}
    \delta^i = r^i - V^i
\end{equation}. They update the value function for the chosen option $i$ in the form of \begin{equation}
    V_{t+1}^{i} = V_{t}^{i} + 
\begin{cases}
    \alpha_C \cdot \delta_{t}^{i}, & \text{if } \delta_{t}^{i} > 0 \\
    \alpha_D \cdot \delta_{t}^{i}, & \text{if } \delta_{t}^{i} < 0
\end{cases}
\end{equation}, and for all unchosen options $i \in \{1,\dots,N\}$ in the form of \begin{equation}
    V_{t+1}^{i} = V_{t}^{i} + 
\begin{cases}
    \alpha_D \cdot \delta_{t}^{i}, & \text{if } \delta_{t}^{i} > 0 \\
    \alpha_C \cdot \delta_{t}^{i}, & \text{if } \delta_{t}^{i} < 0
\end{cases}
\end{equation}, defining there is a confirmatory bias if $\alpha_C > \alpha_D$ and a disconfirmatory bias when $\alpha_C < \alpha_D$. They sample the action based on the probability of $\epsilon$ greedy, a softmax function with a temperature factor or hardmax function on the value function.

\subsection{CM-DQN}
Previous research on integrating confirmation models into reinforcement learning has predominantly focused on discrete state and action spaces \cite{palminteri2023choice}. However, given the inherent continuity of real-world environments, such discretization may not fully capture the complexities of decision-making processes. Deep Q Learning stands as a prominent algorithm within the realm of value-based reinforcement learning \cite{hester2018deep}, offering a robust framework for learning optimal policies. To address the challenge of studying confirmation bias in real-world settings more effectively, we propose a novel deep reinforcement learning algorithm, leveraging neural networks as function approximators to accommodate continuous state inputs. This algorithm, named \textbf{C}onfirmation \textbf{M}odel-based \textbf{D}eep \textbf{Q} \textbf{N}etwork (\textbf{CM-DQN}), extends the applicability of confirmation models to continuous domains.

Nevertheless, we encounter a nuanced dilemma in the optimization of CM-DQN. While gradient descent serves as a ubiquitous tool for minimizing empirical risk, its application in deep reinforcement learning introduces complexities not present in traditional scenarios. Unlike in the context of the multi-armed bandit problem, where adjusting the learning rate directly impacts the updating rule, in gradient descent, simply increasing the learning rate may not necessarily expedite convergence to saddle points. Consequently, in the multi-armed bandit problem, the relative distance between learning rates serves as a proxy for bias, whereas in the realm of deep learning, a supplementary gradient ascent step following gradient descent is employed to emulate the notion of bias in the learning process. We denote the bias type as $B_{\text{bias type}}$ and define as follows:
\begin{equation*}
    B_{\text{bias type}} = \begin{cases}
        B_{\text{confirmatory bias}} \\
        B_{\text{disconfirmatory bias}}  \\
        \text{None}
    \end{cases}
\end{equation*}
\begin{algorithm*}
    \caption{CM-DQN}
    \label{algo: CM-DQN}
    \begin{algorithmic}[1]
    \State Initialize replay buffer with capacity $N$
    \State Initialize action value function network $Q_{\theta}$ and its target action value function network $Q_{\theta_{target}}$with  same random weights
      \For{episode = 1, M}
    \State Initialize sequence $s_1 = \{x_1\}$ and preprocessed sequenced $\phi_1 = \phi(s_1)$
    \For{$t = 1, T$}
        \State With probability $\varepsilon$ select a random action $a_t$
        \State otherwise select $a_t = \max_a Q^*(\phi(s_t), a; \theta)$
        \State Execute action $a_t$ in emulator and observe reward $r_t$ and image $x_{t+1}$
        \State Set $s_{t+1} = s_t, a_t, x_{t+1}$ and preprocess $\phi_{t+1} = \phi(s_{t+1})$
        \State Store transition $(\phi_t, a_t, r_t, \phi_{t+1})$ in $D$
        \State Sample random minibatch of transitions $(\phi_j , a_j , r_j , \phi_{j+1})$ from $D$
        \State Set $y_j =\begin{cases} 
        r_j & \text{for terminal } \phi_{j+1} \\
        r_j + \gamma \max_{a'} Q(\phi_{j+1} , a'; \theta_{target}) & \text{for non-terminal } \phi_{j+11}
        \end{cases}$
        \State \begin{equation*}
            TD_{\text{error}} =y_j – Q(\phi_j ; a_j ; \theta)
        \end{equation*}
        \If{$B_{\text{bias type}}=B_{\text{confirmatory bias}}$ AND $TD_{\text{error}}<0$ OR $B_{\text{bias type}}=B_{\text{disconfirmatory bias}}$ AND  $TD_{\text{error}}>0$}
        \State Perform a gradient descent with step size $\alpha_c$ on $TD_{error}^2$ with respect to $\theta$
        \State Perform a gradient ascent with step size $\alpha_d$ on $TD_{error}^2$ with respect to $\theta$, where $\alpha_d = K\alpha_{c}$ 
        \Else
        \State Perform a gradient descent with step size $\alpha_c$ on $TD_{error}^2$ with respect to $\theta$           \EndIf
    \EndFor
    \State Update $
    \theta_{target} = \tau \theta + (1-\tau) \theta_{target}$

        \State Observe testing reward $r_{\text{test}}$ by doing inference on one more Monte Carlo simulation.
\EndFor
    \end{algorithmic}
\end{algorithm*}
\section{Experiment}
\subsection{Confirmation Model in Multi-Armed Bandit Problem}
Inspired by previous work\cite{lefebvre2022normative}, we consider the confirmation model to model the confirmation bias in the 2-armed bandit problem. We try different pairs of ($\alpha_C$,$\alpha_D$) to explore how the type of confirmation bias and the value of the learning rate affect the average reward one can get.
\subsubsection{Experiment Detail}
\begin{figure}[tp]
\label{figure:ex1}
\begin{center}
\fbox{\includegraphics[width=0.95\linewidth]{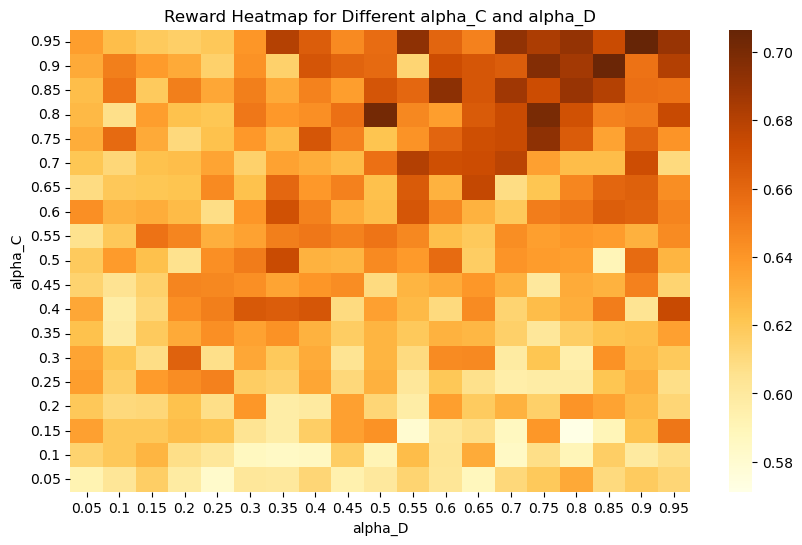}}
\end{center}
\caption{Average Reward for different parameters in 2-armed bandit problem. $\alpha_{C} > \alpha_{D}$ represents the updating rate when there is confirmatory bias and $\alpha_{D} > \alpha_{C}$ stands for the updating rate for disconfirmatory bias.} 
\label{ex1}
\end{figure}

In this work, there are two arms available for selection, each with a distinct stable probability of yielding a reward upon interaction. Specifically, the first arm (arm 0) has a reward probability of $p_1$, and the other has a reward probability of $p_2$, with $p_1$ set to 0.4 and $p_2$ set to 0.6.\\
The rewards are stochastic, employing a binomial distribution where each arm's reward is binary, either a 1 for a reward or a 0 for no reward. The action selection mechanism leverages a \texttt{softmax} function, influenced by the current value estimates of each arm and a temperature parameter that regulates the randomness of the selection process. We set the temperature parameter to be 0.1.\\
Due to time constraints, both $\alpha_C$ and $\alpha_D$ only have the parameter range $\{0.05,0.1,0.15,...,0.90,0.95\}$, and then a grid search is performed for the parameters, totaling 19*19=361 parameter pairs. For each parameter pairs, several trials ($trial number=1024$) are tested and the average reward is the metric for the performance of the model.\\
After running the experiment, the average reward is shown in \autoref{ex1}.

\subsection{CM-DQN in Lunar Lander environment}
The lunar lander environment describes a lander trying to land on a landing pad on the moon. It has the following properties:

\begin{itemize}
\item \textbf{Reward:} The reward, denoted as $r_t$, is a scalar value that reflects the outcome of an agent's action at time step $t$. In the Lunar Lander environment, the reward is typically defined as a combination of factors such as fuel consumption, landing position, and velocity. It is provided by the environment after each action and is used by the agent to learn optimal policies.

\item \textbf{States:} The state of the environment at time step $t$ is represented by a vector $\mathbf{s}_t \in \mathcal{S}$, where $\mathcal{S}$ is the state space. In the Lunar Lander environment, the state vector includes information about the position, velocity, orientation, and angular velocity of the lander, as well as information about the landing pad. 

\item \textbf{Actions:} The action taken by the agent at time step $t$ is denoted as $a_t \in \mathcal{A}$, where $\mathcal{A}$ is the action space. In the Lunar Lander environment, the agent can typically choose from discrete actions such as firing the main engine, firing the side engines, or doing nothing.
\end{itemize}

\subsubsection{Experiment Detail}

In this work, due to the constraint of time, we only search the learning rate among $\{3e-1,3e-2,3e-3,3e-4,3e-5,3e-6\}$ and select $3e-4$ as our learning rate. We present our hyperparameter setting in Table 1. To balance exploration and exploitation, we utilize the $\epsilon$-greedy policy and decrease $\epsilon$ from $0.99$ to $0.01$ as the training episode proceeds.
Inspired by the work \cite{lv2019stochastic}, we add a target Q network to prevent instability during training and update the target network in the form of: \begin{equation*}
    \theta_{target} = \tau \theta + (1-\tau) \theta_{target}
\end{equation*}
After running each episode, we run our experiment on one seed and get the test reward. \autoref{figure 3:experiment result} 3 shows the result of CM-DQN with confirmatory bias, disconfirmatory bias, and without bias.

\begin{table}[H]
\begin{center} 
\caption{Hyperparameters for CM-DQN} 
\label{sample-table} 
\vskip 0.12in
\begin{tabular}{ll} 
\hline
hyperparameter name    &  value \\
\hline
$\tau$        &  5e-2 \\
$\alpha$   &   3e-4 \\
$K$           &   1e-1 \\
$\gamma$        &   0.99 \\
replay buffer size  & 50000 \\
batch size   & 32 \\
Optimizer   & AdamW \\
MLP Dimension & 128 \\
\hline
\end{tabular} 
\end{center} 
\end{table}

\begin{figure}[tp]
\label{figure:Lunar lander}
\begin{center}
\fbox{\includegraphics[width=0.8\linewidth]{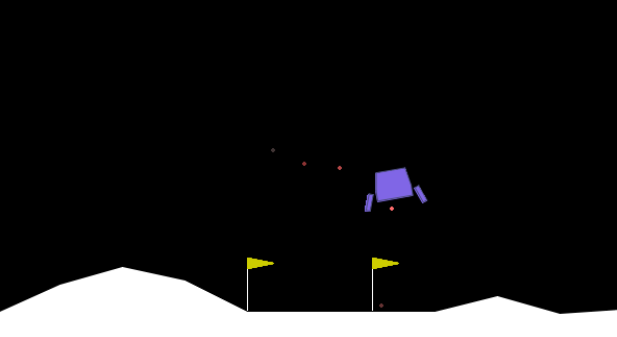}}
\end{center}
\caption{Lunar Lander Environment: the lunar lander tries to land on the surface of moon.} 
\label{sample-figure}
\end{figure}

\begin{figure}[tp]
\label{figure 3:experiment result}
\begin{center}
\fbox{\includegraphics[width=0.95\linewidth]{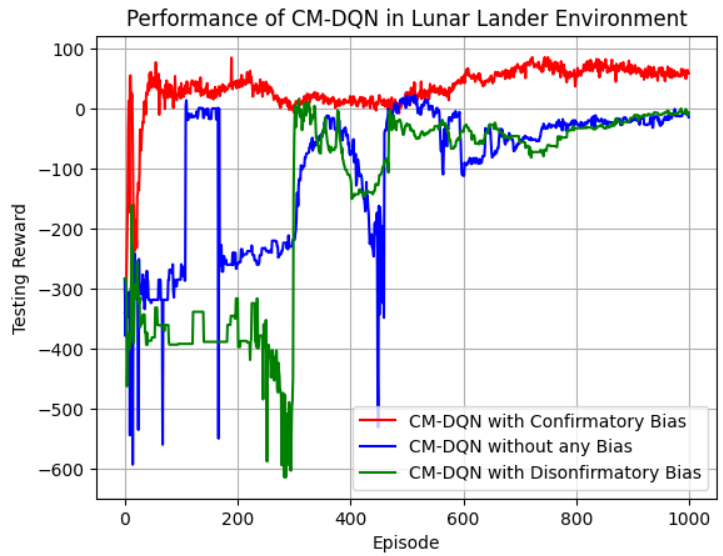}}
\end{center}
\caption{Experiment result of CM-DQN in two types of confirmation bias: X-axis is the episode of training, Y-axis is the testing reward after training in each episode. Confirmatory bias exceeds no bias and disconfirmatory bias.} 
\end{figure}

\begin{figure}[tp]
\begin{center}
\fbox{\includegraphics[width=0.95\linewidth]{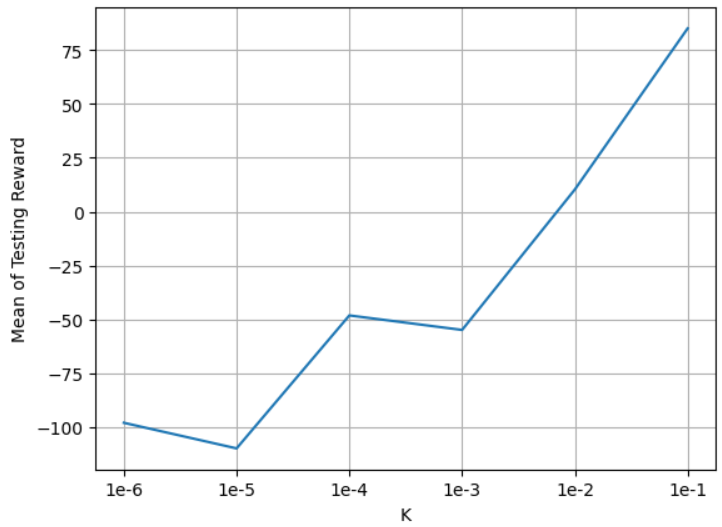}}
\end{center}
\caption{Ablation study of how the bias constraint $K$ impacts on learning outcome of confirmatory bias. The X-axis is the value of $K$. The Y-axis is the averaged testing reward overall episodes after the training process.} 
\label{ablation study}
\end{figure}
\section{Discussion $\And$ Result} 
\subsection{Research Question 1: How does confirmation bias influence decision-making in discrete stationary multi-armed bandit problem?}
For the experiment in the 2-armed bandit problem, we observe that the heatmap (\autoref{ex1}) color gradually darkens from bottom right to top left, indicating that when $\alpha_C$ is larger than $\alpha_D$, the agent tends to learn a better result. As defined in the confirmation model, the observation showcases that agents with confirmatory bias learn better. Besides, the heatmap (\autoref{ex1}) color gradually darkens from bottom left to top right, indicating that as the $\alpha_C$ and $\alpha_D$ increase, the agent tends to learn better as well, which can be a reference when tuning parameters to fit the model.\\
\subsection{Research Question 2: How does confirmation bias influence in continuous state space decision-making process?}
Based on our second experiment of CM-DQN~\ref{algo: CM-DQN} in  \autoref{figure 3:experiment result} 3, the agent learns with confirmatory bias exceeds learning with no bias and disconfirmatory bias in the lunar lander environment. Learning without bias and learning with disconfirmatory bias have similar terminal outcomes around 0. From the view of the result, the agent tends to learn more when the response is consistent with their belief that they will have a better learning outcome in the lunar lander environment. Therefore, we can conclude that confirmatory bias can help the agent gain a larger outcome, while disconfirmatory bias won't significantly influence the learning. 
\section{Ablation Study}
Moreover, given our experiment in Lunar Lander Environment shows CM-DQN with confirmatory bias gains the highest reward, we are curious about how $K$ will influence the learning reward in CM-DQN.
Different from the updating rule of the confirmation model in the multi-armed bandit problem where to play with different learning rates based on the type of belief, in the context of deep learning, we are doing gradient ascent to simulate the "bias" term. We set the $K$ as a constraint to restrict the step size of gradient ascent. However, to explore how the step size can impact the learning reward, we implemented the ablation study of K. We present our result in Figure 4. The result shows $K=1e-1$ has the highest testing reward so we consider using $K=1e-1$ as our bias constraint. By observation, we find out that with larger $K$, the agent trained by CM-DQN with confirmatory bias can gain a higher outcome.
\section{Conclusion}
In this work, we studied the confirmation model in the discrete and continuous state space modeled by the reinforcement learning algorithm. We implemented numerical experiments and concluded that in discrete and continuous state space, agents with confirmatory bias get the highest award. With the CM-DQN model, more tasks with continuous states and discrete actions can be explored and the corresponding human decision-making behaviors can be tested and modeled, which can help the understanding of confirmation bias from a cognitive science perspective.\\
However, we didn't average the result over more random seeds due to the time limit, so some randomness may still exist in the experiment results.
Future work about fitting \textbf{CM-DQN} in more decision-making tasks and observing human behaviors in continuous states and discrete actions can be conducted. In terms of algorithmic level, integrating the confirmation model into Deep Deterministic Policy Gradient to study confirmation bias in continuous state and continuous action decision processes could also be further explored.
\section{Acknowledgments}
We are thankful to Professor Brenden Lake and Professor Todd Gureckis for providing us with valuable feedback and instructions. We also appreciate it for NYU Greene Cluster provides high-performance computing resources.

\nocite{ChalnickBillman1988a}
\nocite{Feigenbaum1963a}
\nocite{Hill1983a}
\nocite{OhlssonLangley1985a}
\nocite{Matlock2001}
\nocite{NewellSimon1972a}
\nocite{ShragerLangley1990a}

\bibliographystyle{apacite}

\setlength{\bibleftmargin}{.125in}
\setlength{\bibindent}{-\bibleftmargin}

\bibliography{main}

\end{document}